\documentclass[letterpaper]{article} 
\usepackage[preprint]{aaai2027}  
\usepackage[hyphens]{url}  
\usepackage{graphicx} 
\usepackage{natbib}  
\usepackage{caption} 
\usepackage{algorithm}
\usepackage{algorithmic}

\usepackage{tabularx}
\usepackage{amssymb}
\usepackage{amsmath}
\usepackage{subcaption}

\usepackage{array}
\usepackage{pifont}

\newcolumntype{Y}{>{\centering\arraybackslash}X}
\newcolumntype{C}[1]{>{\centering\arraybackslash}p{#1}}
\usepackage{newfloat}
\usepackage{listings}
\DeclareCaptionStyle{ruled}{labelfont=normalfont,labelsep=colon,strut=off} 
\floatstyle{ruled}
\newfloat{listing}{tb}{lst}{}
\floatname{listing}{Listing}

\usepackage{booktabs}

\title{DHCG: Dynamic Construction of Hierarchical Collaboration Graphs for LLM-Based Multi-Agent Reasoning}
\author{
Jie Ren\textsuperscript{\rm 1},
Jiakang Yuan\textsuperscript{\rm 1},
Chenyu Huang\textsuperscript{\rm 1},
Hezeer Ma\textsuperscript{\rm 1},
Jiayuan Fan\textsuperscript{\rm 1},
Tao Chen\textsuperscript{\rm 1, 2}\thanks{Corresponding Author}
}
\affiliations{
    \textsuperscript{\rm 1}College of Future Information Technology, Fudan University, Shanghai, China \\
    \textsuperscript{\rm 2}Shanghai Innovation Institute, Shanghai, China \\

}

\begin{document}

\maketitle

\begin{abstract}
LLM-based multi-agent systems (MAS) have demonstrated strong capabilities
in solving complex problems across diverse domains. Recently, the dynamic orchestration of agent systems has become an important research direction.
However, existing methods suffer from limited composition, misaligned dependencies, and inflexible scale, restricting their ability to adapt to reasoning requirements during execution. To address these limitations, we reframe MAS
design as a partially observable Markov decision process, in which
both the composition and scale of the MAS are dynamically
determined. We propose DHCG, a novel framework that coordinates three
modules (Planner, Worker, and Generator) to progressively construct a dynamic
hierarchical collaboration graph from scratch based on the query and
evolving execution feedback. At each step, guided by feedback, the Planner
generates a set of distinct and complementary roles tailored to the current
reasoning needs and selectively routes relevant information to each role. It can also finalize the hierarchical collaboration graph early or progressively expand it when additional reasoning is required.
We further introduce action-aware preference optimization to train the
Planner to make more effective decisions when constructing hierarchical
collaboration graphs. We systematically evaluate DHCG across code generation, mathematical reasoning, and domain-specific reasoning benchmarks. DHCG achieves state-of-the-art average performance among the compared methods, improving over the single-agent baseline by \textbf{13.06} points and outperforming both static and dynamic MAS baselines by \textbf{2.77--8.02} points. Additional experiments further demonstrate its generalization across different Planner backbones and unseen Worker models.
\end{abstract}


\section{Introduction}

Agents powered by large language models (LLMs) have demonstrated strong capabilities across a wide range of tasks, including code generation, decision making, and question answering~\citep{react,reflexion,codeagent,toolformer}.
Despite these advances, a single agent may still face challenges when solving complex problems involving diverse expertise and verification from multiple perspectives.
Recent studies have increasingly explored LLM-based multi-agent systems (MAS)~\cite{autogen,team2025internagent,camel}, where multiple agents collaborate through role specialization and communication. Early frameworks such as AutoGen~\citep{autogen}, CAMEL~\citep{camel}, and MetaGPT~\citep{metagpt} largely rely on manually designed roles and interaction patterns, requiring considerable human effort and providing limited task-specific adaptability. These limitations have motivated growing efforts to automatically design
and adapt multi-agent collaboration for different tasks.

\begin{figure}
    \centering
    \makebox[\columnwidth][c]{%
        \includegraphics[width=1.00\columnwidth]{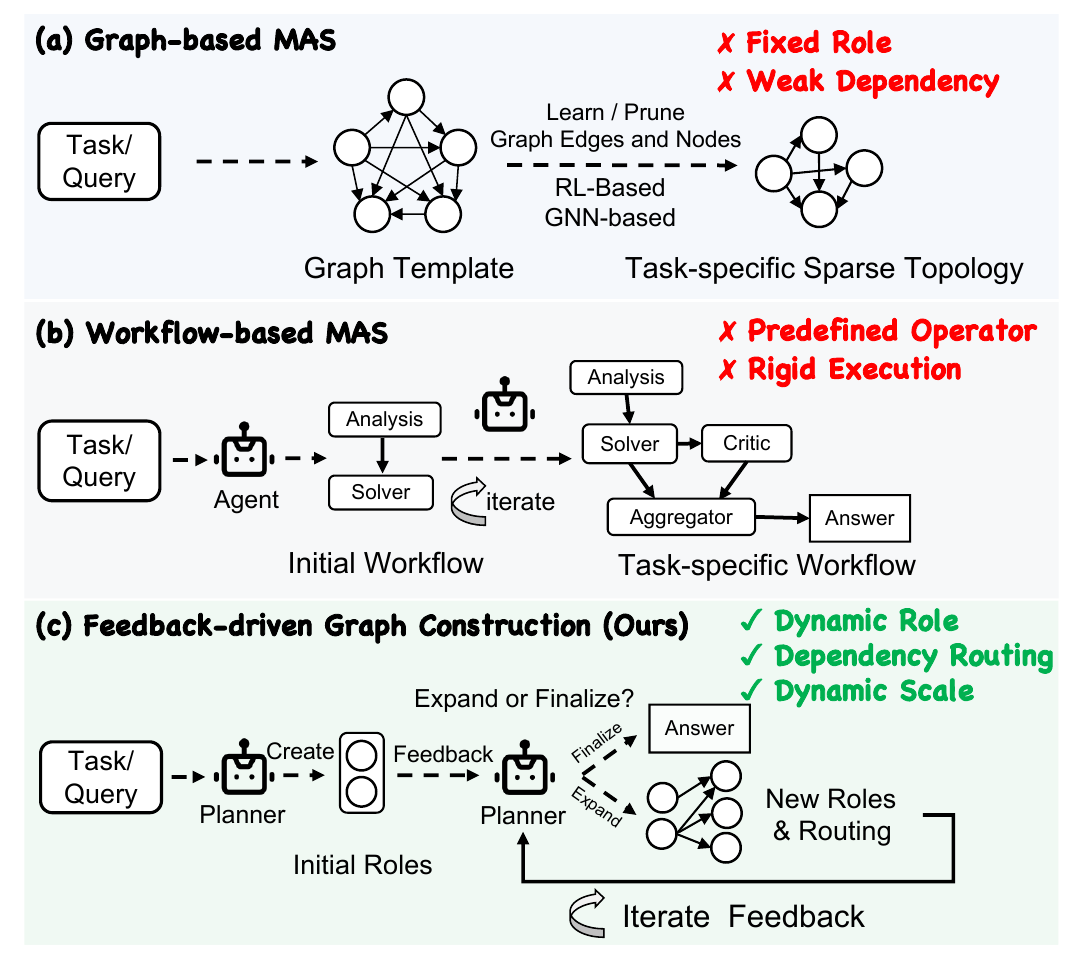}
    }
    \caption{
    Comparison of three MAS construction paradigms:
    (a) graph-based MAS;
    (b) workflow-based MAS; and
    (c) our feedback-driven graph construction paradigm.
    }
    \label{fig:motivation}
\end{figure}

Existing approaches to automated MAS design can be broadly categorized into
two paradigms: graph-based MAS and workflow-based MAS, as shown in
Fig.~\ref{fig:motivation}(a) and Fig.~\ref{fig:motivation}(b). Graph-based MAS optimize inter-agent communication topologies through graph optimization, pruning, or topology generation~\citep{gptswarm,agentprune,li2026assemble}.
A representative example is G-Designer~\citep{zhang2025gdesigner}, which employs graph neural networks to generate a task-adaptive communication topology. In parallel, workflow-based MAS construct and optimize executable workflows by organizing components through search, evolution, or feedback-driven refinement~\citep{adas,evomas,mas2}. For example, AFlow~\citep{zhang2025aflow} uses Monte Carlo tree search to iteratively optimize workflows.

Despite their task-adaptive designs, existing dynamic MAS methods still
suffer from several limitations.
\textbf{\textit{1) Limited Composition.}}
Existing methods construct MAS from predefined agent nodes or workflow
operators.
Consequently, they may provide insufficient role coverage because the agent
composition cannot adapt to task-specific requirements or emerging reasoning
needs during execution.
\textbf{\textit{2) Misaligned Dependencies.}}
Communication dependencies are often determined by predefined templates or global
optimization objectives, rather than the actual information needs
of downstream agents. Agents may receive irrelevant or redundant
context while missing critical intermediate evidence, which can weaken
collaboration and propagate errors through the reasoning process. \textbf{\textit{3) Inflexible scale.}}
Fixed interaction rounds and one-shot workflow generation determine the
reasoning scale before execution.
This can cause the system to continue reasoning after sufficient evidence has
been obtained, or stop before the remaining reasoning needs have been
adequately addressed.
Given the above limitations, a natural question arises:
\textit{Can we dynamically design MAS at inference time, adapting their
components, dependencies, and scale to the evolving reasoning state?}

To seek the answer to the above question, we draw inspiration from how real-world
teams incrementally recruit experts according to evolving task needs,
rather than assembling all members in advance.
This motivates a \textit{feedback-driven dynamic collaboration graph
construction} paradigm for MAS design, as shown in
Fig.~\ref{fig:motivation}(c).
Unlike existing paradigms that determine agent components, communication
structures, and interaction scales before execution, our paradigm
constructs the MAS progressively from scratch using real-time execution
feedback. At each construction step, unresolved reasoning
needs are identified, and an appropriate number of complementary roles are
generated, enabling task- and state-specific capability composition.
Dependencies are then established according to each new role's information
requirements, aligning communication with downstream needs.
Finally, the construction process adaptively determines whether to expand the
collaboration graph or finalize the reasoning process, allowing the system scale
to evolve with the current reasoning state. Together, these mechanisms help alleviate the limitations in role composition, dependency alignment, and system scalability, respectively.

Building on this new paradigm, we propose \textbf{DHCG}:
\textbf{D}ynamic construction of \textbf{H}ierarchical
\textbf{C}ollaboration \textbf{G}raphs for MAS. We formulate MAS construction
as a partially observable Markov decision process, in which a Planner makes
sequential decisions based on the query and evolving execution feedback. At each layer, the Planner generates complementary roles for unresolved reasoning needs, assigns relevant predecessors through dependency-aware instructions, and decides whether to expand or finalize the graph. Consequently, both the
collaboration structure and scale evolve with the reasoning process. Training such a sequential construction policy online is expensive. We therefore devise an offline action-aware preference
optimization strategy that reuses collected trajectory trees to construct
multiple preference pairs from relatively few trajectories. It evaluates alternative Planner actions through downstream outcomes,
constructs within- and cross-action preferences, and weights
the pairs by their value gaps. This offline training strategy guides the Planner toward more effective role-generation, dependency-routing, and finalization decisions. Empirical results show that DHCG achieves the highest average performance,
outperforming the strongest static, workflow-based, and graph-based MAS
baselines by 4.76, 2.77, and 3.33 points, respectively. The framework also
generalizes across Planner backbones and transfers effectively to unseen
Worker models. Additional analyses show that DHCG adapts its collaboration
structure and scale to task difficulty.

Our main contributions are summarized as follows:

\begin{itemize}
    \item We introduce an inference-time dynamic graph construction paradigm
    for MAS design, where collaboration structures are progressively constructed
    according to the query and real-time execution feedback, rather than being
    fully specified before execution.

    \item We propose \textbf{DHCG}, a Planner-driven framework that formulates
    MAS construction as a POMDP and adaptively determines roles, dependencies,
    and graph scale. We further introduce offline action-aware preference
    optimization, which derives within- and cross-action preferences from
    downstream trajectory quality and action-level execution cost.

    \item We evaluate DHCG on seven benchmarks spanning
    mathematical reasoning, code generation, and general knowledge. The results
    validate DHCG's effectiveness and its generalization across Planner backbones
    and unseen Worker models. Additional analyses show that DHCG adapts its
    collaboration structure and scale to task difficulty.
\end{itemize}

\section{Related Work}
\label{sec:related_work}

\paragraph{Static and Dynamic MAS Construction.}
Early MAS frameworks organize collaboration through predefined roles and
protocols. CAMEL~\citep{camel} and AutoGen~\citep{autogen} support structured
agent interactions, while MetaGPT~\citep{metagpt} coordinates specialized
agents through standardized operating procedures. Such static designs rely heavily on manual specification
and offer limited adaptability across tasks. Recent dynamic MAS methods seek to replace manually specified collaboration
structures with automatically optimized designs. At the topology level, GPTSwarm~\citep{gptswarm} and
G-Designer~\citep{zhang2025gdesigner} learn task-adaptive communication structures,
while AgentPrune~\citep{agentprune} and
AgentDropout~\citep{wang2025agentdropout} prune redundant agents or edges.
At the workflow level, ADAS~\citep{adas} and AFlow~\citep{zhang2025aflow} search for
executable procedures, whereas MASS~\citep{mass} and MaAS~\citep{zhang2025maas}
optimize broader agent configurations and orchestration. These methods reduce manual design and improve task-specific coordination, but predefined components or search spaces still limit adaptation during execution.

\paragraph{Policy Optimization for MAS Construction.}
Recent studies increasingly formulate MAS construction as a learnable
decision-making process. One line of work optimizes policies for generating
complete workflows or global collaboration structures.
ScoreFlow~\citep{scoreflow} learns workflow generation from
execution-score-derived preferences, while MAS$^2$~\citep{mas2} combines
collaborative tree search with value-guided optimization to train multiple
Meta-Agents. AgentConductor~\citep{agentconductor} further introduces
difficulty-aware rewards for optimizing layered communication topologies.
Another line of work focuses on sequential construction guided by intermediate
feedback. DyFlow~\citep{dyflow} trains a Designer through distilled self-play
to adapt its planning decisions, whereas Workflow-R1~\citep{workflowr1} and
FlowSteer~\citep{flowsteer} optimize multi-turn workflow construction through
sub-sequence policy updates and atomic workflow editing, respectively.
Collectively, these studies demonstrate the effectiveness of policy
optimization for automated MAS construction. However, the former provides only coarse supervision at the workflow or
structure level, while the latter focuses on sequential refinement and offers
limited modeling of evolving inter-agent dependencies.

\begin{figure*}[t]
    \centering
    \includegraphics[width=0.94\textwidth]{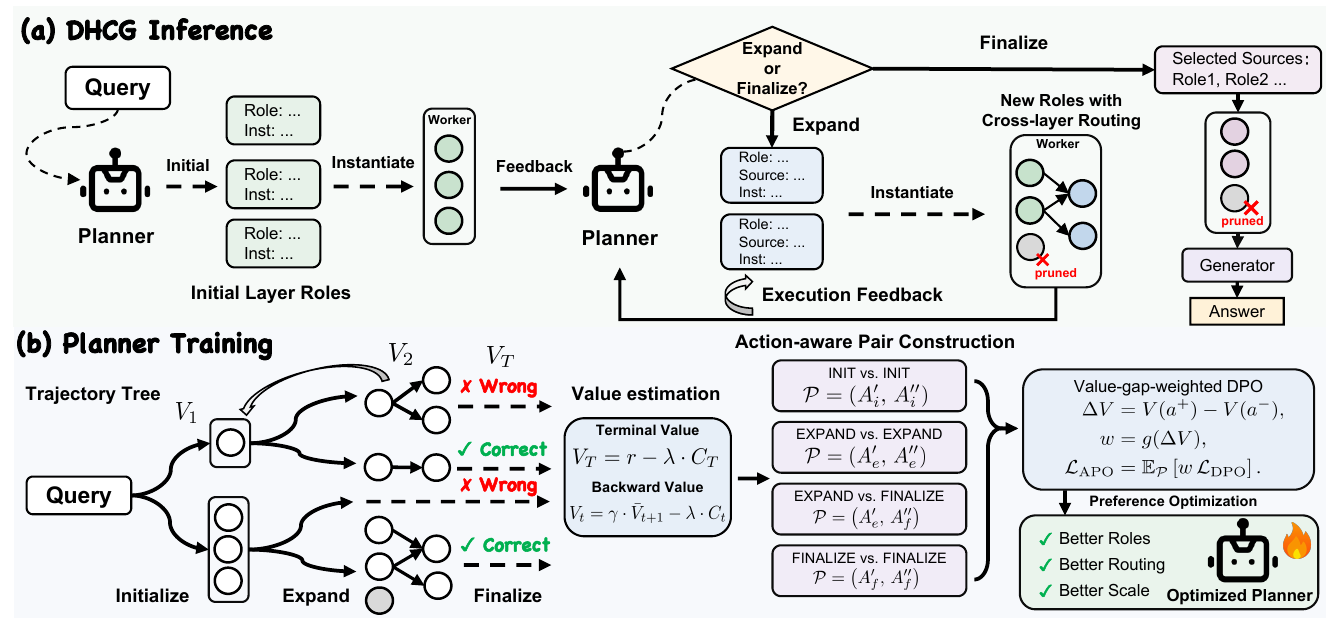}
\caption{
Overview of DHCG.
(a) The Planner dynamically initializes, expands, and finalizes the hierarchical collaboration graph during inference.
(b) The Planner is trained from tree-structured trajectories via action-aware preference optimization.
}
    \label{fig:overview}
\end{figure*}

\section{Methodology}
\subsection{Problem Formulation}

\paragraph{Hierarchical Collaboration Graph.}
We model multi-agent collaboration as a hierarchical directed acyclic graph \(G=(V,E)\).
Each node \(v_i\in V\) denotes an agent execution, and each edge \((v_j,v_i)\in E\) routes the output of \(v_j\) to \(v_i\), thereby defining an information dependency.
Agent \(v_i\) is specified by a language model policy \(\pi_{\theta_i}\) and an instruction prompt \(p_i\).
Given the original task \(q\), its routed context and output are defined as
\begin{equation}
x_i=\{y_j\mid(v_j,v_i)\in E\},
\qquad
y_i\sim\pi_{\theta_i}(\cdot\mid q,p_i,x_i),
\end{equation}
where \(x_i\) collects the outputs \(y_j\) routed from the predecessors of \(v_i\), and \(y_i\) denotes the output generated by \(v_i\). For a node without incoming edges, \(x_i=\emptyset\).

To distinguish sequential dependencies across reasoning stages from parallel
collaboration within the same stage, we partition \(G\) into layers:
\begin{equation}
V=\bigsqcup_{l=0}^{L}V_l,
\qquad
E\subseteq\bigcup_{l=1}^{L}
\left(V_{l-1}\times V_l\right),
\end{equation}
where $V_l$ denotes nodes in layer $l$, $L$ is the maximum layer index,
and $E$ denotes inter-layer edges. This constraint excludes intra-layer
dependencies, allowing agents in the same layer to execute in parallel.

\paragraph{Dynamic Graph Construction as a POMDP.}
Rather than determining all agent nodes and dependencies before execution, we formulate collaboration graph construction as a sequential decision process in which a planner adapts the graph according to intermediate agent outputs.

We model this process as a partially observable Markov decision process
(POMDP):
\begin{equation}
\mathcal{M}
=
\langle
\mathcal{S},\mathcal{A},T,R,\Omega,O,\gamma
\rangle,
\end{equation}
where \(\mathcal{S}\) is the latent state space of graph construction,
\(\mathcal{A}\) is the Planner action space, \(T\) is the transition function,
\(\Omega\) is the observation space, and \(O\) is the observation function.
The reward function \(R\) reflects downstream task performance and execution
cost, while \(\gamma\) is the discount factor.

At decision step \(t\), the latent state \(s_t\in\mathcal{S}\) summarizes
the graph-construction history, accumulated agent outputs, and unobserved
reasoning conditions.
The Planner observes the query and the role-output pairs from the most recently
executed layer:
\begin{equation}
o_t=O(s_t)=
\begin{cases}
(q,\emptyset), & t=0,\\
(q,\mathcal{Z}_{t-1}), & t\geq 1,
\end{cases}
\qquad o_t\in\Omega,
\end{equation}
where
\(\mathcal{Z}_{t-1}=\{(\rho_i,y_i)\mid v_i\in V_{t-1}\}\), with
\(\rho_i\) and \(y_i\) denoting the role specification and output of agent
node \(v_i\).

Conditioned on \(o_t\), the planner selects an action and transitions to the next state:
\begin{equation}
a_t\sim\pi_{\theta}(\cdot\mid o_t),
\qquad
s_{t+1}\sim T(\cdot\mid s_t,a_t),
\end{equation}
where \(\pi_{\theta}\) is the planner policy, \(a_t\) is the selected construction action, and \(T\) describes the transition induced by the action and subsequent agent execution.

\subsection{Dynamic Hierarchical Collaboration Graph}
\paragraph{Overview.}
Building on the POMDP formulation, we propose \textbf{DHCG}, a framework that dynamically constructs hierarchical collaboration graphs for MAS. As shown in Figure~\ref{fig:overview}(a), DHCG consists of a Planner, Workers, and a Generator. Based on the query and evolving execution feedback, the Planner incrementally generates task-specific roles, specifies inter-layer dependencies, and decides whether to expand or finalize the graph. Workers execute the generated roles to produce intermediate outputs, while the Generator synthesizes the final answer from the outputs selected by the Planner. This design enables parallel intra-layer collaboration and hierarchical inter-layer information flow.

\paragraph{Initial Layer Construction.} The initial layer provides a set of independent and complementary perspectives for addressing the input query. Since communication dependencies may propagate and amplify errors introduced at early reasoning stages \citep{shen2025understanding}, DHCG avoids premature information coupling by allowing the initial agents to reason independently. Meanwhile, the required perspectives can vary across queries. Rather than relying on fixed agent configurations or constructing structures solely from learned query representations (e.g., embeddings) \citep{su2026difficulty}, DHCG employs a Planner to analyze the task-specific reasoning requirements of each query. Based on this analysis, the Planner determines both the number of initial agents and the complementary roles. 

Formally, given the input query \(q\), the Planner generates the initial layer as
\begin{equation}
V_0 = \{v_i=(\rho_i,p_i)\}_{i=1}^{n_0}
\sim \pi_{\theta}^{\mathrm{init}}(\cdot \mid q),
\end{equation}
where \(n_0\) denotes the number of agents, while \(\rho_i\) and \(p_i\) denote the role specification and instruction prompt of agent \(v_i\), respectively. The corresponding Workers are instantiated from these nodes and execute their assigned roles in parallel. Let \(y_i^{(0)}\) denote the output produced by the Worker associated with node \(v_i\) in the initial layer. The resulting role-output pairs are collected as
\begin{equation}
\mathcal{Z}_0=\{(\rho_i,y_i^{(0)})\}_{i=1}^{n_0},
\end{equation}
which are passed to the Planner for subsequent planning steps.

\paragraph{Dynamic Planning and Graph Construction.} After executing the initial layer, DHCG continues graph construction based on the latest execution feedback. Fixed interaction rounds or execution stages may lead to unnecessary computation for simple queries or insufficient reasoning for complex ones \citep{wang2025agentdropout}. Moreover, dense or predefined communication structures may introduce redundant information or fail to provide the evidence required by downstream agents~\citep{qian2025scaling}. Accordingly, DHCG adaptively determines whether to finalize or expand the graph and, when expanding, establishes sparse inter-layer dependencies by routing relevant outputs from the preceding layer to each newly generated role. 

Formally, after layer \(V_{t-1}\) is executed, the Planner observes the query \(q\) and the latest role-output pairs \(\mathcal{Z}_{t-1}\), and samples a dynamic planning action:
\begin{equation}
a_t \sim \pi_{\theta}^{\mathrm{dyn}}(\cdot \mid q,\mathcal{Z}_{t-1}),
\qquad a_t \in \mathcal{A}_t,
\end{equation}
where the action space is defined as
\begin{equation}
\mathcal{A}_t =
\left\{
\begin{array}{ll}
\mathrm{FINALIZE}(C_{t-1}), & C_{t-1}\subseteq V_{t-1}, \\[2pt]
\mathrm{EXPAND}(V_t,E_t), & E_t\subseteq V_{t-1}\times V_t .
\end{array}
\right.
\end{equation}
If the Planner selects \(\mathrm{FINALIZE}(C_{t-1})\), the construction process is terminated and the selected source agents are used for final answer synthesis. Otherwise, if the Planner selects \(\mathrm{EXPAND}(V_t,E_t)\), it generates a new layer of agents:
\begin{equation}
V_t
=
\left\{
v_i^{(t)}
=
\bigl(\rho_i^{(t)},p_i^{(t)}\bigr)
\right\}_{i=1}^{n_t},
\end{equation}
where \(n_t\) denotes the number of newly generated agents in layer \(t\).
For each new agent \(v_i^{(t)}\), the Planner selects only the necessary
source agents from the previous layer, denoted by
\(\mathcal{S}_i^{(t)} \subseteq V_{t-1}\).
Agents whose outputs are irrelevant, redundant, or unnecessary for
\(v_i^{(t)}\) are not connected. The corresponding sparse routing edges are then constructed as
\begin{equation}
E_t=\{(v_j,v_i^{(t)}) \mid v_i^{(t)}\in V_t,\; v_j\in \mathcal{S}_i^{(t)}\}.
\end{equation}
The instruction prompt $p_i^{(t)}$ specifies both the new agent's role and how it should process the routed source outputs, such as by verifying, refining, comparing, or synthesizing them. The corresponding Workers are then instantiated to execute these roles, producing role--output pairs. These role--output pairs are returned to the Planner for the next decision step, enabling DHCG to progressively adapt the collaboration graph according to the evolving intermediate results.

\paragraph{Answer Synthesis.}
DHCG employs a dedicated Generator to produce a stable and well-formatted
final response. Upon \(\mathrm{FINALIZE}(C_{t-1})\), only the selected
role-output pairs in \(C_{t-1}\) are provided to the Generator; if the maximum
layer \(L\) is reached, all outputs in \(V_L\) are used instead. This design
filters redundant context and enables focused answer synthesis.

\subsection{Planner Training}

The Planner jointly determines construction actions, agent roles, and routing dependencies, which directly affect both the collaboration graph and final answer quality. Inspired by tree-based preference construction~\citep{mas2}, we propose \textbf{Action-aware Preference Optimization} to train the Planner from the downstream outcomes of alternative construction decisions. As shown in Figure~\ref{fig:overview}(b), we first collect tree-structured trajectories by recursively sampling and executing alternative Planner actions. We then construct action-aware preference pairs and optimize the Planner according to their relative advantages.

\paragraph{Action-aware Preference Construction.} Given a planning observation $o_t$, we sample multiple candidate Planner
outputs and recursively execute them to construct a tree of alternative
graph-construction trajectories. Each node $n_t$ represents a candidate
Planner action at construction step $t$.

We recursively estimate the value of each candidate action as
\begin{equation}
V_t
=
\begin{cases}
r-\lambda \cdot C_T, & t=T,\\[1mm]
\gamma \cdot \bar{V}_{t+1}-\lambda \cdot C_t, & t<T,
\end{cases}
\label{eq:action_aware_value}
\end{equation}
where $r\in[0,1]$ denotes the final-answer score, $C_t$ is the token
cost incurred by the current Planner action, $\lambda$ controls the
trade-off between accuracy and efficiency, and $\gamma\in(0,1]$ is the
discount factor. Here,
$\bar{V}_{t+1}$ denotes the average recursively estimated value of the
directly sampled child actions.

We construct preferences based on the action types being compared. Cross-action pairs, primarily \textsc{Expand}--\textsc{Finalize}, use task-only values to avoid favoring \textsc{Finalize} due to lower cost, whereas same-action pairs use cost-aware values to account for execution efficiency. We retain only pairs with sufficiently large value gaps and discard same-action pairs when both candidates have zero downstream task scores.

\begin{algorithm}[t]
\caption{DHCG Inference}
\label{alg:dhcg}
\begin{algorithmic}[1]
\REQUIRE Query $q$; maximum layer $L$; width budget $W$
\ENSURE Final answer $\hat{y}$ and collaboration graph $G$

\STATE \textbf{Step 1: Initial Layer Construction}
\STATE $V_0\leftarrow\mathrm{Planner}_{\mathrm{init}}(q),
       \quad |V_0|\leq W,
       \quad G\leftarrow(V_0,\emptyset)$
\STATE $\mathcal{Z}_0\leftarrow\mathrm{Workers}(q,V_0),
       \quad C\leftarrow V_0$

\STATE \textbf{Step 2: Dynamic Graph Construction}
\FOR{$t=1$ to $L$}
    \STATE $a_t\leftarrow
    \mathrm{Planner}_{\mathrm{dyn}}(q,\mathcal{Z}_{t-1})$
    \IF{$a_t=\mathrm{FINALIZE}(C_{t-1}),
         \quad C_{t-1}\subseteq V_{t-1}$}
        \STATE $C\leftarrow C_{t-1}$; \textbf{break}
    \ENDIF
    \STATE $(V_t,E_t)\leftarrow a_t,
           \quad |V_t|\leq W,
           \quad G\leftarrow G\cup(V_t,E_t)$
    \STATE $\mathcal{Z}_t\leftarrow
    \mathrm{Workers}(q,V_t,E_t,\mathcal{Z}_{t-1}),
    \quad C\leftarrow V_t$
\ENDFOR

\STATE \textbf{Step 3: Answer Synthesis}
\STATE $\hat{y}\leftarrow
\mathrm{Generator}(q,\mathrm{Outputs}(C))$
\STATE \textbf{return} $(\hat{y},G)$
\end{algorithmic}
\end{algorithm}

\newcommand{\gain}[1]{%
  \nobreak\hspace{0.08em}%
  {\fontsize{5.5}{6}\selectfont$\uparrow\!#1$}}
\newcommand{\drop}[1]{%
  \nobreak\hspace{0.08em}%
  {\fontsize{5.5}{6}\selectfont$\downarrow\!#1$}}

\begin{table*}[t]
\centering

\small
\renewcommand{\arraystretch}{1.12}
\setlength{\tabcolsep}{2pt}

\begin{tabularx}{1.0\textwidth}{@{}l|*{7}{Y}|Y@{}}
\toprule
\textbf{Method}
& \textbf{MBPP}
& \textbf{HumanEval}
& \textbf{MATH-500}
& \textbf{AIME}
& \textbf{GSM8K}
& \mbox{\textbf{GPQA}}
& \textbf{MMLU-Pro}
& \textbf{Average} \\
\midrule

IO
& 63.73
& 82.70
& 81.20
& 17.78
& 88.09
& 37.54
& 51.57
& 60.37 \\

CoT
& 70.48\gain{6.75}
& 85.75\gain{3.05}
& 84.60\gain{3.40}
& 18.89\gain{1.11}
& 92.58\gain{4.49}
& 39.90\gain{2.36}
& 54.29\gain{2.72}
& 63.78\gain{3.41} \\

CoT-SC
& 71.36\gain{7.63}
& 86.01\gain{3.31}
& 88.47\gain{7.27}
& 25.56\gain{7.78}
& 93.90\gain{5.81}
& 52.02\gain{14.48}
& 64.95\gain{13.38}
& 68.90\gain{8.52} \\

\midrule

LLM-Debate
& 71.07\gain{7.34}
& 88.04\gain{5.34}
& \underline{88.60}\gain{7.40}
& 23.33\gain{5.55}
& \textbf{94.19}\gain{6.10}
& 52.53\gain{14.99}
& 62.95\gain{11.38}
& 68.67\gain{8.30} \\

MacNet
& 68.33\gain{4.60}
& 84.48\gain{1.78}
& 86.00\gain{4.80}
& 26.67\gain{8.89}
& 91.53\gain{3.44}
& 49.66\gain{12.12}
& 63.57\gain{12.00}
& 67.18\gain{6.80} \\

\midrule

G-Designer
& 78.59\gain{14.86}
& 85.49\gain{2.79}
& 87.80\gain{6.60}
& 25.56\gain{7.78}
& 93.37\gain{5.28}
& \underline{53.54}\gain{16.00}
& \underline{66.38}\gain{14.81}
& 70.10\gain{9.73} \\

ARG-Designer
& 72.63\gain{8.90}
& 86.26\gain{3.56}
& 83.60\gain{2.40}
& 20.00\gain{2.22}
& 91.34\gain{3.25}
& 53.37\gain{15.83}
& 64.81\gain{13.24}
& 67.43\gain{7.06} \\

\midrule

AFlow
& 72.33\gain{8.60}
& 84.48\gain{1.78}
& 87.07\gain{5.87}
& 21.11\gain{3.33}
& 92.86\gain{4.77}
& 43.10\gain{5.56}
& 56.95\gain{5.38}
& 65.41\gain{5.04} \\

MaAS
& \underline{82.01}\gain{18.28}
& \underline{89.82}\gain{7.12}
& 87.33\gain{6.13}
& \underline{28.89}\gain{11.11}
& 92.89\gain{4.80}
& 49.83\gain{12.29}
& 63.86\gain{12.29}
& 70.66\gain{10.29} \\

\midrule

\textbf{DHCG (Ours)}
& \textbf{84.36}\gain{20.63}
& \textbf{93.13}\gain{10.43}
& \textbf{88.73}\gain{7.53}
& \textbf{30.00}\gain{12.22}
& \underline{93.93}\gain{5.84}
& \textbf{55.89}\gain{18.35}
& \textbf{68.00}\gain{16.43}
& \textbf{73.43}\gain{13.06} \\

\bottomrule
\end{tabularx}

\caption{Performance comparison (\%) on seven benchmarks.}
\label{tab:main_results}

\end{table*}

\paragraph{Action-aware Preference Optimization.}
Each retained preference pair is represented as
$(o_i,a_i^{+},a_i^{-},w_i)$, where $w_i$ is obtained by clipping the
corresponding value gap. We use a smaller clipping threshold for same-action
pairs, which are more frequent and partially reflect efficiency differences.
The resulting weights are normalized by their mean over the training set,
yielding $\widetilde{w}_i$.

Let $\pi_{\mathrm{ref}}$ denote the frozen reference Planner. For the $i$-th
preference pair, we define the relative log-likelihood difference as
\begin{equation}
\Delta\rho_{\theta}^{(i)}
=
\log
\frac{\pi_{\theta}(a_i^{+}\mid o_i)}
     {\pi_{\mathrm{ref}}(a_i^{+}\mid o_i)}
-
\log
\frac{\pi_{\theta}(a_i^{-}\mid o_i)}
     {\pi_{\mathrm{ref}}(a_i^{-}\mid o_i)}.
\label{eq:relative_log_likelihood_gap}
\end{equation}

The Planner is optimized using the weighted DPO for
action-aware preference optimization:
\begin{equation}
\mathcal{L}_{\mathrm{APO}}
=
-\frac{1}{N}
\sum_{i=1}^{N}
\widetilde{w}_i
\log \sigma
\left(
\beta\Delta\rho_{\theta}^{(i)}
\right),
\label{eq:weighted_dpo}
\end{equation}
where $\sigma$ is the sigmoid function and $\beta$ controls the strength of
preference separation relative to the reference Planner. 
Cross-action preferences teach the Planner whether further graph expansion is beneficial or the current graph should be finalized, whereas same-action preferences help it distinguish among alternative decisions of the same type by considering both downstream task performance and execution efficiency.

\section{Experiments}

\subsection{Experiment Setup}
\paragraph{Benchmarks.}
To comprehensively evaluate DHCG, we conduct experiments on seven benchmarks
spanning three reasoning domains.
For code reasoning, we use HumanEval~\citep{chen2021humaneval} and
MBPP~\citep{austin2021mbpp}.
For mathematical reasoning, we adopt
MATH-500~\citep{lightman2024verify} to assess general
mathematical problem solving, GSM8K~\citep{cobbe2021gsm8k} to evaluate
multi-step arithmetic reasoning, and AIME~2025~\citep{maa2025aime} to probe
performance on challenging competition-level problems.
For knowledge-intensive reasoning, we use
GPQA Diamond~\citep{rein2024gpqa} to evaluate graduate-level scientific
reasoning and MMLU-Pro~\citep{wang2024mmlupro} to assess complex discipline reasoning.

\paragraph{Baselines.}

We compare DHCG with representative baselines from four categories.
Single-model prompting includes IO (direct LLM invocation),
Chain-of-Thought~\citep{wei2022cot}, and Self-Consistency CoT (5 answers)~\citep{wang2023selfconsistency}.
Static MAS include LLM-Debate~\citep{llmdebate} and
MacNet~\citep{qian2025scaling}.
Graph-based MAS include G-Designer~\citep{zhang2025gdesigner} and
ARG-Designer~\citep{li2026assemble}. Workflow-based MAS include AFlow~\citep{zhang2025aflow} and
MaAS~\citep{zhang2025maas}.
For a fair comparison, all agent types in every method use the same LLM
backbone and identical inference configurations. Details of benchmark curation, data splits, baseline configurations, and
evaluation protocols are provided in Appendix A.
\paragraph{Implementation Details.}
Our Planner is fine-tuned from Qwen3-8B~\citep{yang2025qwen3}, while both the Workers and Generator use the original
Qwen3-8B model. All experiments are conducted over three
independent runs, and we report the average performance.
For Planner training, we sample the training data from
DeepMath-103K~\citep{he2025deepmath} and
KodCode~\citep{xu2025kodcode}. Details of the training-data composition,
selection and preprocessing procedures, preference-data construction, and
optimization configurations are provided in
Appendix B.

\subsection{Experimental Results}
\paragraph{Overall Performance.}
Table~\ref{tab:main_results} reports the results across seven benchmarks.
DHCG achieves the highest average accuracy of 73.43\%, outperforming the strongest static MAS method, LLM-Debate, by 4.76 points, the strongest graph-based method, G-Designer, by 3.33 points, and the strongest workflow-based method, MaAS, by 2.77 points.
Among the baselines, MaAS performs particularly well on code-generation
tasks, while G-Designer is strong on knowledge-intensive reasoning.
DHCG achieves strong performance across code generation, mathematical
reasoning, and knowledge-intensive reasoning, demonstrating the
effectiveness of dynamically constructing collaboration graphs according
to evolving reasoning states. Notably, although the Planner is trained
only on mathematical and code-generation data, DHCG remains highly
competitive on GPQA and MMLU-Pro, indicating effective cross-domain
generalization. This generalization is supported by DHCG's dynamic graph
construction, which adapts roles and dependencies to evolving reasoning
needs.

\paragraph{Empirical Upper Bound.}
Table~\ref{tab:pass5} reports the Pass@5 results of DHCG and CoT under repeated sampling. DHCG consistently outperforms CoT across all three benchmarks, indicating a higher probability of obtaining a correct solution over multiple runs. This advantage stems from the diverse collaboration graphs and reasoning paths constructed across different runs. In particular, DHCG achieves a Pass@5 of 98.47\% on HumanEval, demonstrating a strong empirical upper bound for code generation. The improvements on MATH-500 and MMLU-Pro further show that this advantage extends to mathematical reasoning and multi-domain knowledge reasoning.

\begin{table}[t]
\centering

\small
\renewcommand{\arraystretch}{1.12}
\setlength{\tabcolsep}{3pt}

\begin{tabularx}{\columnwidth}{@{}
>{\raggedright\arraybackslash}p{0.14\columnwidth}|
*{3}{>{\centering\arraybackslash}X}
|>{\centering\arraybackslash}X@{}}
\toprule
\textbf{Method}
& \textbf{HumanEval}
& \textbf{MATH-500}
& \mbox{\textbf{MMLU-Pro}}
& \textbf{Avg.} \\
\midrule

CoT
& 92.37
& 93.40
& 74.68
& 86.82 \\

\textbf{DHCG}
& \textbf{98.47}
& \textbf{95.20}
& \textbf{81.71}
& \textbf{91.79} \\

\bottomrule
\end{tabularx}

\caption{Pass@5 performance (\%) on three representative benchmarks.}
\label{tab:pass5}

\end{table}

\subsection{Generalization Analysis}

\paragraph{Cross-Planner Generalization.}
To assess the robustness of DHCG to different Planner backbones, we instantiate
the framework with three distinct models: GPT-5.4-mini~\citep{singh2025openai},
DeepSeek-V4-Pro~\citep{deepseekai2026deepseekv4}, and our trained Qwen3-8B Planner. As shown in
Table~\ref{tab:cross_planner}, all three variants achieve higher average
accuracy than the strongest baseline, MaAS, indicating that DHCG is not tied
to a specific Planner model and generalizes well across different Planner
families. Moreover, although our Planner is trained from the smaller
8B backbone, it achieves an average score comparable to those achieved
by larger Planner models. These results
suggest that our training strategy can effectively
elicit the planning capability of a compact model, enabling competitive
performance without relying on large-scale Planner models.

\begin{table}[t]
\centering

\small
\renewcommand{\arraystretch}{1.12}
\setlength{\tabcolsep}{2pt}

\begin{tabularx}{\columnwidth}{@{}
>{\raggedright\arraybackslash}p{0.35\columnwidth}|
*{3}{>{\centering\arraybackslash}X}
>{\centering\arraybackslash}X@{}}
\toprule
\textbf{Method}
& \textbf{MBPP}
& \textbf{AIME}
& \mbox{\textbf{MMLU-Pro}}
& \textbf{Avg.} \\
\midrule

MaAS
& 82.01
& 28.89
& 63.86
& 58.25 \\

\midrule

DHCG (GPT-5.4-mini)
& 84.16
& 28.89
& 69.24
& 60.76 \\

DHCG (DS-V4-Pro)
& 83.97
& 28.89
& \textbf{69.48}
& 60.78 \\

DHCG (Qwen3-8B$^\dagger$)
& \textbf{84.36}
& \textbf{30.00}
& 68.00
& \textbf{60.79} \\

\bottomrule
\end{tabularx}

\caption{Cross-Planner generalization of DHCG and comparison with baselines
(\%) on three benchmarks. $^\dagger$ denotes our Planner trained from the
Qwen3-8B backbone.}
\label{tab:cross_planner}

\end{table}

\begin{figure}
    \centering
\includegraphics[width=0.95\columnwidth]{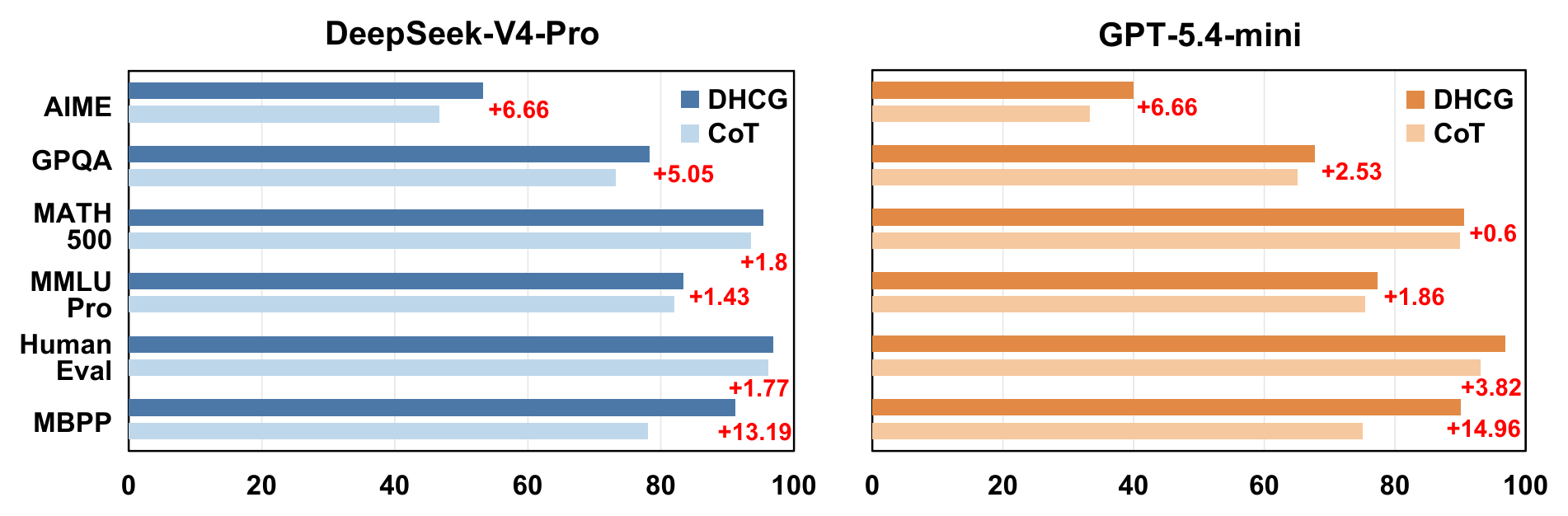}
\caption{
Cross-worker generalization of DHCG against the corresponding single-model
CoT baselines.
}
    \label{fig:cross_worker}
\end{figure}

\paragraph{Cross-Worker Generalization.}
We further investigate whether DHCG is sensitive to the choice of Worker
model. To this end, we keep the trained Planner fixed and pair it with
DeepSeek-V4-Pro and GPT-5.4-mini as Workers, respectively. As shown in
Figure~\ref{fig:cross_worker}, DHCG consistently outperforms the corresponding
single-model CoT baseline across the evaluated benchmarks, although the
magnitude of improvement varies by task. These results demonstrate that the
learned graph-construction policy can effectively generalize across different
Worker backbones, improving reasoning quality while maintaining robust and
consistent performance.

\begin{table}[t]
\centering

\small
\renewcommand{\arraystretch}{1.12}
\setlength{\tabcolsep}{1.8pt}

\begin{tabularx}{\columnwidth}{@{}
>{\raggedright\arraybackslash}p{0.28\columnwidth}|
>{\centering\arraybackslash}p{0.19\columnwidth}
*{3}{>{\centering\arraybackslash}X}@{}}
\toprule
\textbf{Method}
& \mbox{\textbf{MMLU-Pro}}
& \textbf{GSM8K}
& \textbf{HumanEval}
& \textbf{Avg.} \\
\midrule

\multicolumn{5}{@{}l}{\textit{Planner Training}} \\

Base Planner
& 67.14
& 93.55
& 91.35
& 84.01 \\

SFT
& 66.29
& 93.81
& 88.30
& 82.80 \\

Vanilla DPO
& 66.71
& 93.40
& 92.37
& 84.16 \\

\midrule

\multicolumn{5}{@{}l}{\textit{Graph Construction}} \\

w/o Dyn. Routing
& 67.00
& 93.71
& 91.86
& 84.19\\

w/o Dyn. Scale
& 65.62
& 92.76
& 88.80
& 82.39\\

w/o Dyn. Role
& 66.19
& 93.78
& 89.31
& 83.09\\

\midrule

\textbf{DHCG}
& \textbf{68.00}
& \textbf{93.93}
& \textbf{93.13}
& \textbf{85.02} \\

\bottomrule
\end{tabularx}

\caption{Ablation results (\%) on three representative benchmarks.}
\label{tab:ablation}

\end{table}

\subsection{Ablation Study}

\paragraph{Planner Training.}
We compare the base Planner, SFT, vanilla DPO, and our action-aware preference
optimization. As shown in Table~\ref{tab:ablation}, SFT reduces the average
performance by 1.21, including a 3.05 drop on
HumanEval. Since SFT optimizes token-level imitation, the Planner may overfit to specific
planning patterns, reducing its ability to adapt
its decisions to evolving execution feedback. This observation also supports our choice
to directly optimize the base Planner through reinforcement
learning, rather than adopting the conventional SFT-then-RL training
paradigm. Vanilla DPO uses the same value-based preference pairs as our method but
assigns a uniform weight to every pair. Its slight improvement over the
base Planner confirms the effectiveness of our preference construction.
Our complete method further weights each pair according to its value gap and
achieves the best average score, demonstrating the additional benefit
of distinguishing preference strength during optimization.

\paragraph{Graph Construction.}
We ablate dynamic routing, graph-scale control, and dynamic role generation
by replacing them with fixed alternatives. Removing any of these dynamic components leads to performance degradation to
varying degrees. Among them, dynamic graph-scale control has the largest overall impact, with its removal reducing average performance by 2.63 and HumanEval performance by 4.33. This highlights the necessity of adapting
the collaboration scale to evolving reasoning requirements rather than
determining it in advance. Together, these results show that dynamic scale,
role generation, and routing provide complementary benefits for constructing
effective collaboration graphs.

\subsection{Framework Analysis}

\begin{figure}[t]
    \centering
    \includegraphics[width=0.95\columnwidth]{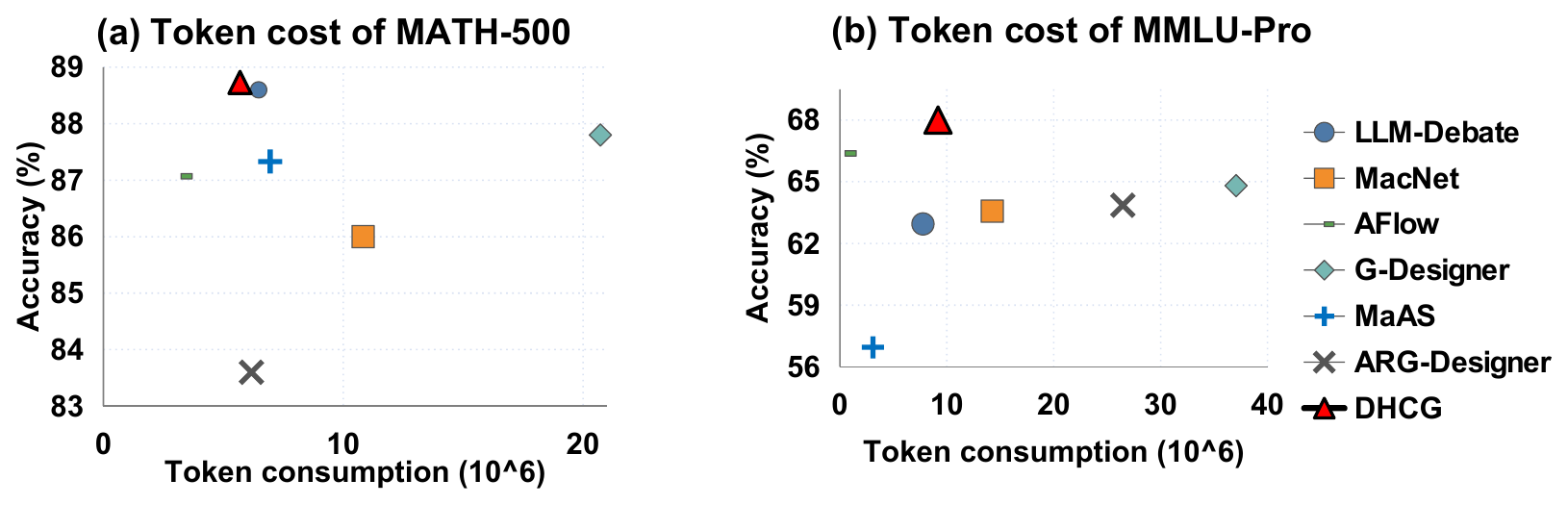}
    \caption{
    Performance and total token cost comparison.
    }
    \label{fig:token_cost}
\end{figure}

\paragraph{Cost Analysis.}
DHCG dynamically adapts the collaboration graph to task-specific reasoning requirements, allocating different amounts of computational resources across queries. As shown in Figure~\ref{fig:token_cost}, fixed graph templates and static MAS methods can introduce substantial redundant communication. This advantage is particularly evident on MMLU-Pro, where DHCG outperforms G-Designer with over 50\% fewer tokens and achieves higher accuracy than MacNet while reducing token consumption by 32.16\%. A similar accuracy--efficiency advantage is observed on MATH-500, where DHCG surpasses MaAS with comparable token usage, whereas AFlow consumes fewer tokens but achieves lower accuracy. Overall, these results demonstrate that dynamic graph construction improves the balance between reasoning performance and computational cost.

\paragraph{Difficulty-Aware Graph Construction.}
We further examine whether DHCG adapts its collaboration graph to problem difficulty. Both the MATH-500 trend analysis and qualitative case study show that DHCG constructs larger graphs with more Workers for harder problems, while maintaining compact structures for easier ones. These results demonstrate that DHCG dynamically allocates computational resources according to reasoning difficulty. Additional quantitative and qualitative analyses are provided in Appendix C.

\section{Conclusion}

In this work, we proposed DHCG, a novel framework for the dynamic
construction of hierarchical collaboration graphs in MAS. We formulated MAS construction
as a partially observable Markov decision process and introduced a new paradigm in which
the collaboration graph is progressively constructed during execution. DHCG employs a Planner to generate task-specific roles, determine dependency routing, and control graph scale through sequential expansion and finalization decisions based on evolving execution feedback, while optimizing the planning policy through action-aware preference optimization.
Extensive experiments demonstrate DHCG's effectiveness
and generalization across Planner backbones and unseen Worker models,
with our compact Planner performing comparably to larger ones.

\section*{Acknowledgement}
This work is supported by the National Key R\&D Program of China (No. 2026YFE0101200), and the National Natural Science Foundation of China (No. 62671175). The computations in this research were performed using the CFFF platform of Fudan University.

\bibliography{aaai2027}

@inproceedings{camel,
  title     = {{CAMEL}: Communicative Agents for ``Mind'' Exploration of Large Language Model Society},
  author    = {Li, Guohao and Hammoud, Hasan Abed Al Kader and Itani, Hani and Khizbullin, Dmitrii and Ghanem, Bernard},
  booktitle = {Advances in Neural Information Processing Systems},
  volume    = {36},
  year      = {2023},
  doi       = {10.52202/075280-2264},
  url       = {https://proceedings.neurips.cc/paper/2023/hash/a3621ee907def47c1b952ade25c67698-Abstract-Conference.html}
}

@inproceedings{autogen,
  title     = {{AutoGen}: Enabling Next-Gen {LLM} Applications via Multi-Agent Conversations},
  author    = {Wu, Qingyun and Bansal, Gagan and Zhang, Jieyu and Wu, Yiran and Li, Beibin and Zhu, Erkang and Jiang, Li and Zhang, Xiaoyun and Zhang, Shaokun and Liu, Jiale and Awadallah, Ahmed Hassan and White, Ryen W. and Burger, Doug and Wang, Chi},
  booktitle = {Proceedings of the First Conference on Language Modeling},
  year      = {2024},
  url       = {https://openreview.net/forum?id=BAakY1hNKS}
}

@inproceedings{metagpt,
  title     = {{MetaGPT}: Meta Programming for a Multi-Agent Collaborative Framework},
  author    = {Hong, Sirui and Zhuge, Mingchen and Chen, Jonathan and Zheng, Xiawu and Cheng, Yuheng and Zhang, Ceyao and Wang, Jinlin and Wang, Zili and Yau, Steven Ka Shing and Lin, Zijuan and Zhou, Liyang and Ran, Chenyu and Xiao, Lingfeng and Wu, Chenglin and Schmidhuber, J{\"u}rgen},
  booktitle = {The Twelfth International Conference on Learning Representations},
  year      = {2024},
  url       = {https://openreview.net/forum?id=VtmBAGCN7o}
}

@inproceedings{gptswarm,
  title     = {{GPTSwarm}: Language Agents as Optimizable Graphs},
  author    = {Zhuge, Mingchen and Wang, Wenyi and Kirsch, Louis and Faccio, Francesco and Khizbullin, Dmitrii and Schmidhuber, J{\"u}rgen},
  booktitle = {Proceedings of the 41st International Conference on Machine Learning},
  series    = {Proceedings of Machine Learning Research},
  volume    = {235},
  pages     = {62743--62767},
  publisher = {PMLR},
  year      = {2024},
  url       = {https://proceedings.mlr.press/v235/zhuge24a.html}
}

@inproceedings{agentprune,
  title     = {Cut the Crap: An Economical Communication Pipeline for {LLM}-based Multi-Agent Systems},
  author    = {Zhang, Guibin and Yue, Yanwei and Li, Zhixun and Yun, Sukwon and Wan, Guancheng and Wang, Kun and Cheng, Dawei and Yu, Jeffrey Xu and Chen, Tianlong},
  booktitle = {The Thirteenth International Conference on Learning Representations},
  year      = {2025},
  eprint    = {2410.02506},
  archivePrefix = {arXiv},
  primaryClass  = {cs.MA},
  url       = {https://arxiv.org/abs/2410.02506}
}

@inproceedings{wang2025agentdropout,
  title     = {{AgentDropout}: Dynamic Agent Elimination for Token-Efficient and High-Performance {LLM}-Based Multi-Agent Collaboration},
  author    = {Wang, Zhexuan and Wang, Yutong and Liu, Xuebo and Ding, Liang and Zhang, Miao and Liu, Jie and Zhang, Min},
  booktitle = {Proceedings of the 63rd Annual Meeting of the Association for Computational Linguistics (Volume 1: Long Papers)},
  pages     = {24013--24035},
  publisher = {Association for Computational Linguistics},
  address   = {Vienna, Austria},
  year      = {2025},
  doi       = {10.18653/v1/2025.acl-long.1170},
  url       = {https://aclanthology.org/2025.acl-long.1170/}
}

@inproceedings{zhang2025gdesigner,
  title     = {{G-Designer}: Architecting Multi-Agent Communication Topologies via Graph Neural Networks},
  author    = {Zhang, Guibin and Yue, Yanwei and Sun, Xiangguo and Wan, Guancheng and Yu, Miao and Fang, Junfeng and Wang, Kun and Chen, Tianlong and Cheng, Dawei},
  booktitle = {Proceedings of the 42nd International Conference on Machine Learning},
  series    = {Proceedings of Machine Learning Research},
  volume    = {267},
  pages     = {76678--76692},
  publisher = {PMLR},
  year      = {2025},
  url       = {https://proceedings.mlr.press/v267/zhang25cu.html}
}

@article{li2026assemble,
  title   = {Assemble Your Crew: Automatic Multi-Agent Communication Topology Design via Autoregressive Graph Generation},
  author  = {Li, Shiyuan and Liu, Yixin and Wen, Qingsong and Zhang, Chengqi and Pan, Shirui},
  journal = {Proceedings of the AAAI Conference on Artificial Intelligence},
  volume  = {40},
  number  = {28},
  pages   = {23142--23150},
  year    = {2026},
  doi     = {10.1609/aaai.v40i28.39481},
  url     = {https://ojs.aaai.org/index.php/AAAI/article/view/39481}
}

@inproceedings{adas,
  title     = {Automated Design of Agentic Systems},
  author    = {Hu, Shengran and Lu, Cong and Clune, Jeff},
  booktitle = {The Thirteenth International Conference on Learning Representations},
  year      = {2025},
  eprint    = {2408.08435},
  archivePrefix = {arXiv},
  primaryClass  = {cs.AI},
  url       = {https://proceedings.iclr.cc/paper_files/paper/2025/hash/36b7acf6f6010652b3f2a433774a66fe-Abstract-Conference.html}
}

@inproceedings{zhang2025aflow,
  title     = {{AFlow}: Automating Agentic Workflow Generation},
  author    = {Zhang, Jiayi and Xiang, Jinyu and Yu, Zhaoyang and Teng, Fengwei and Chen, Xionghui and Chen, Jiaqi and Zhuge, Mingchen and Cheng, Xin and Hong, Sirui and Wang, Jinlin and Zheng, Bingnan and Liu, Bang and Luo, Yuyu and Wu, Chenglin},
  booktitle = {The Thirteenth International Conference on Learning Representations},
  year      = {2025},
  eprint    = {2410.10762},
  archivePrefix = {arXiv},
  primaryClass  = {cs.AI},
  url       = {https://arxiv.org/abs/2410.10762}
}

@inproceedings{mass,
  title     = {Multi-Agent Design: Optimizing Agents with Better Prompts and Topologies},
  author    = {Zhou, Han and Wan, Xingchen and Sun, Ruoxi and Palangi, Hamid and Iqbal, Shariq and Vuli{\'c}, Ivan and Korhonen, Anna and Ar{\i}k, Sercan {\"O}.},
  booktitle = {The Fourteenth International Conference on Learning Representations},
  year      = {2026},
  eprint    = {2502.02533},
  archivePrefix = {arXiv},
  primaryClass  = {cs.LG},
    url = {https://openreview.net/forum?id=I05H9RUzHB}
}

@inproceedings{zhang2025maas,
  title     = {Multi-Agent Architecture Search via Agentic Supernet},
  author    = {Zhang, Guibin and Niu, Luyang and Fang, Junfeng and Wang, Kun and Bai, Lei and Wang, Xiang},
  booktitle = {Proceedings of the 42nd International Conference on Machine Learning},
  series    = {Proceedings of Machine Learning Research},
  volume    = {267},
  pages     = {75834--75852},
  publisher = {PMLR},
  year      = {2025},
  url       = {https://proceedings.mlr.press/v267/zhang25bi.html}
}

@article{scoreflow,
  title         = {{ScoreFlow}: Mastering {LLM} Agent Workflows via Score-based Preference Optimization},
  author        = {Wang, Yinjie and Yang, Ling and Li, Guohao and Wang, Mengdi and Aragam, Bryon},
  journal       = {arXiv preprint arXiv:2502.04306},
  year          = {2025},
  eprint        = {2502.04306},
  archivePrefix = {arXiv},
  primaryClass  = {cs.CL},
  doi           = {10.48550/arXiv.2502.04306},
  url           = {https://arxiv.org/abs/2502.04306}
}

@article{mas2,
  title         = {{MAS}$^2$: Self-Generative, Self-Configuring, Self-Rectifying Multi-Agent Systems},
  author        = {Wang, Kun and Zhang, Guibin and Ye, ManKit and Deng, Xinyu and Wang, Dongxia and Hu, Xiaobin and Guo, Jinyang and Liu, Yang and Guo, Yufei},
  journal       = {arXiv preprint arXiv:2509.24323},
  year          = {2025},
  eprint        = {2509.24323},
  archivePrefix = {arXiv},
  primaryClass  = {cs.MA},
  doi           = {10.48550/arXiv.2509.24323},
  url           = {https://arxiv.org/abs/2509.24323}
}

@article{agentconductor,
  title         = {{AgentConductor}: Topology Evolution for Multi-Agent Competition-Level Code Generation},
  author        = {Wang, Siyu and Lu, Ruotian and Yang, Zhihao and Wang, Yuchao and Zhang, Yanzhou and Xu, Lei and Xu, Qimin and Yin, Guojun and Chen, Cailian and Guan, Xinping},
  journal       = {arXiv preprint arXiv:2602.17100},
  year          = {2026},
  eprint        = {2602.17100},
  archivePrefix = {arXiv},
  primaryClass  = {cs.MA},
  doi           = {10.48550/arXiv.2602.17100},
  url           = {https://arxiv.org/abs/2602.17100}
}

@article{dyflow,
  title         = {{DyFlow}: Dynamic Workflow Framework for Agentic Reasoning},
  author        = {Wang, Yanbo and Xu, Zixiang and Huang, Yue and Wang, Xiangqi and Song, Zirui and Gao, Lang and Wang, Chenxi and Tang, Xiangru and Zhao, Yue and Cohan, Arman and Zhang, Xiangliang and Chen, Xiuying},
  journal       = {arXiv preprint arXiv:2509.26062},
  year          = {2025},
  eprint        = {2509.26062},
  archivePrefix = {arXiv},
  primaryClass  = {cs.CL},
  doi           = {10.48550/arXiv.2509.26062},
  url           = {https://arxiv.org/abs/2509.26062}
}

@article{workflowr1,
  title         = {{Workflow-R1}: Group Sub-Sequence Policy Optimization for Multi-Turn Workflow Construction},
  author        = {Kong, Mingze and Qu, Zikun and Zhou, Zhongquan and Liang, Pengyu and Li, Xiang and Shang, Zhiwei and Hong, Zhi and Huang, Kaiyu and Wang, Zhiyong and Dai, Zhongxiang},
  journal       = {arXiv preprint arXiv:2602.01202},
  year          = {2026},
  eprint        = {2602.01202},
  archivePrefix = {arXiv},
  primaryClass  = {cs.AI},
  doi           = {10.48550/arXiv.2602.01202},
  url           = {https://arxiv.org/abs/2602.01202}
}

@article{flowsteer,
  title         = {{FlowSteer}: Towards Agents Designing Agentic Workflows via Reinforced Progressive Canvas Editing},
  author        = {Zhang, Mingda and Liu, Wenjin and Shen, Tiesunlong and Lin, Qika and Mao, Rui and Cambria, Erik and Tang, Xiaoying and Luo, Haoran},
  journal       = {arXiv preprint arXiv:2602.01664},
  year          = {2026},
  eprint        = {2602.01664},
  archivePrefix = {arXiv},
  primaryClass  = {cs.AI},
  doi           = {10.48550/arXiv.2602.01664},
  url           = {https://arxiv.org/abs/2602.01664}
}

@inproceedings{shen2025understanding,
  title     = {Understanding the Information Propagation Effects of Communication Topologies in {LLM}-Based Multi-Agent Systems},
  author    = {Shen, Xu and Liu, Yixin and Dai, Yiwei and Wang, Yili and Miao, Rui and Tan, Yue and Pan, Shirui and Wang, Xin},
  booktitle = {Proceedings of the 2025 Conference on Empirical Methods in Natural Language Processing},
  pages     = {12347--12361},
  publisher = {Association for Computational Linguistics},
  address   = {Suzhou, China},
  year      = {2025},
  doi       = {10.18653/v1/2025.emnlp-main.623},
  url       = {https://aclanthology.org/2025.emnlp-main.623/}
}

@inproceedings{su2026difficulty,
  title     = {Difficulty-Aware Agentic Orchestration for Query-Specific Multi-Agent Workflows},
  author    = {Su, Jinwei and Lan, Qizhen and Xia, Yinghui and Sun, Lifan and Tian, Weiyou and Shi, Tianyu and He, Lewei},
  booktitle = {Proceedings of the ACM Web Conference 2026},
  pages     = {2060--2070},
  publisher = {ACM},
  year      = {2026},
  doi       = {10.1145/3774904.3792240},
  url       = {https://doi.org/10.1145/3774904.3792240}
}

@inproceedings{qian2025scaling,
  title     = {Scaling Large Language Model-Based Multi-Agent Collaboration},
  author    = {Qian, Chen and Xie, Zihao and Wang, YiFei and Liu, Wei and Zhu, Kunlun and Xia, Hanchen and Dang, Yufan and Du, Zhuoyun and Chen, Weize and Yang, Cheng and Liu, Zhiyuan and Sun, Maosong},
  booktitle = {The Thirteenth International Conference on Learning Representations},
  year      = {2025},
  url       = {https://proceedings.iclr.cc/paper_files/paper/2025/hash/66a026c0d17040889b50f0dfa650e5e0-Abstract-Conference.html}
}

@article{evomas,
  title         = {{EvoMAS}: Evolutionary Generation of Multi-Agent Systems},
  author        = {Hu, Yuntong and Zhang, Yuting and Trager, Matthew and Zhang, Yi and Yang, Shuo and Xia, Wei and Soatto, Stefano},
  journal       = {arXiv preprint arXiv:2602.06511},
  year          = {2026},
  eprint        = {2602.06511},
  archivePrefix = {arXiv},
  primaryClass  = {cs.LG},
  doi           = {10.48550/arXiv.2602.06511},
  url           = {https://arxiv.org/abs/2602.06511}
}

@inproceedings{react,
  title     = {{ReAct}: Synergizing Reasoning and Acting in Language Models},
  author    = {Yao, Shunyu and Zhao, Jeffrey and Yu, Dian and Du, Nan and Shafran, Izhak and Narasimhan, Karthik and Cao, Yuan},
  booktitle = {The Eleventh International Conference on Learning Representations},
  year      = {2023},
  url       = {https://openreview.net/forum?id=WE_vluYUL-X}
}

@inproceedings{reflexion,
  title     = {Reflexion: Language Agents with Verbal Reinforcement Learning},
  author    = {Shinn, Noah and Cassano, Federico and Gopinath, Ashwin and Narasimhan, Karthik and Yao, Shunyu},
  booktitle = {Advances in Neural Information Processing Systems},
  volume    = {36},
  year      = {2023},
  doi       = {10.52202/075280-0377},
  url       = {https://proceedings.neurips.cc/paper/2023/hash/1b44b878bb782e6954cd888628510e90-Abstract-Conference.html}
}

@article{codeagent,
  title         = {{CodeAgent}: Enhancing Code Generation with Tool-Integrated Agent Systems for Real-World Repo-level Coding Challenges},
  author        = {Zhang, Kechi and Li, Jia and Li, Ge and Shi, Xianjie and Jin, Zhi},
  journal       = {arXiv preprint arXiv:2401.07339},
  year          = {2024},
  eprint        = {2401.07339},
  archivePrefix = {arXiv},
  primaryClass  = {cs.SE},
  doi           = {10.48550/arXiv.2401.07339},
  url           = {https://arxiv.org/abs/2401.07339}
}

@inproceedings{toolformer,
  title     = {{Toolformer}: Language Models Can Teach Themselves to Use Tools},
  author    = {Schick, Timo and Dwivedi-Yu, Jane and Dess{\`i}, Roberto and Raileanu, Roberta and Lomeli, Maria and Hambro, Eric and Zettlemoyer, Luke and Cancedda, Nicola and Scialom, Thomas},
  booktitle = {Advances in Neural Information Processing Systems},
  volume    = {36},
  year      = {2023},
  url       = {https://proceedings.neurips.cc/paper/2023/hash/d842425e4bf79ba039352da0f658a906-Abstract-Conference.html}
}

@article{chen2021humaneval,
  title         = {Evaluating Large Language Models Trained on Code},
  author        = {Chen, Mark and Tworek, Jerry and Jun, Heewoo and others},
  journal       = {arXiv preprint arXiv:2107.03374},
  year          = {2021},
  eprint        = {2107.03374},
  archivePrefix = {arXiv},
  primaryClass  = {cs.LG},
  doi           = {10.48550/arXiv.2107.03374},
  url           = {https://arxiv.org/abs/2107.03374}
}

@article{austin2021mbpp,
  title         = {Program Synthesis with Large Language Models},
  author        = {Austin, Jacob and Odena, Augustus and Nye, Maxwell and Bosma, Maarten and Michalewski, Henryk and Dohan, David and Jiang, Ellen and Cai, Carrie and Terry, Michael and Le, Quoc V. and Sutton, Charles},
  journal       = {arXiv preprint arXiv:2108.07732},
  year          = {2021},
  eprint        = {2108.07732},
  archivePrefix = {arXiv},
  primaryClass  = {cs.PL},
  doi           = {10.48550/arXiv.2108.07732},
  url           = {https://arxiv.org/abs/2108.07732}
}

@inproceedings{lightman2024verify,
  title     = {Let's Verify Step by Step},
  author    = {Lightman, Hunter and Kosaraju, Vineet and Burda, Yura and Edwards, Harri and Baker, Bowen and Lee, Teddy and Leike, Jan and Schulman, John and Sutskever, Ilya and Cobbe, Karl},
  booktitle = {The Twelfth International Conference on Learning Representations},
  year      = {2024},
  url       = {https://openreview.net/forum?id=v8L0pN6EOi}
}

@article{cobbe2021gsm8k,
  title         = {Training Verifiers to Solve Math Word Problems},
  author        = {Cobbe, Karl and Kosaraju, Vineet and Bavarian, Mohammad and Chen, Mark and Jun, Heewoo and Kaiser, Lukasz and Plappert, Matthias and Tworek, Jerry and Hilton, Jacob and Nakano, Reiichiro and Hesse, Christopher and Schulman, John},
  journal       = {arXiv preprint arXiv:2110.14168},
  year          = {2021},
  eprint        = {2110.14168},
  archivePrefix = {arXiv},
  primaryClass  = {cs.LG},
  doi           = {10.48550/arXiv.2110.14168},
  url           = {https://arxiv.org/abs/2110.14168}
}

@misc{maa2025aime,
  title        = {American Invitational Mathematics Examination ({AIME})},
  author       = {{Mathematical Association of America}},
  howpublished = {MAA Invitational Competitions},
  year         = {2025},
  url          = {https://maa.org/maa-invitational-competitions/}
}

@inproceedings{rein2024gpqa,
  title     = {{GPQA}: A Graduate-Level Google-Proof {Q\&A} Benchmark},
  author    = {Rein, David and Hou, Betty Li and Stickland, Asa Cooper and Petty, Jackson and Pang, Richard Yuanzhe and Dirani, Julien and Michael, Julian and Bowman, Samuel R.},
  booktitle = {Proceedings of the First Conference on Language Modeling},
  year      = {2024},
  url       = {https://openreview.net/forum?id=Ti67584b98}
}

@inproceedings{wang2024mmlupro,
  title     = {{MMLU-Pro}: A More Robust and Challenging Multi-Task Language Understanding Benchmark},
  author    = {Wang, Yubo and Ma, Xueguang and Zhang, Ge and Ni, Yuansheng and Chandra, Abhranil and Guo, Shiguang and Ren, Weiming and Arulraj, Aaran and He, Xuan and Jiang, Ziyan and Li, Tianle and Ku, Max and Wang, Kai and Zhuang, Alex and Fan, Rongqi and Yue, Xiang and Chen, Wenhu},
  booktitle = {Advances in Neural Information Processing Systems},
  volume    = {37},
  year      = {2024},
  doi       = {10.52202/079017-3018},
  url       = {https://proceedings.neurips.cc/paper_files/paper/2024/hash/ad236edc564f3e3156e1b2feafb99a24-Abstract.html}
}

@inproceedings{wei2022cot,
  title     = {Chain-of-Thought Prompting Elicits Reasoning in Large Language Models},
  author    = {Wei, Jason and Wang, Xuezhi and Schuurmans, Dale and Bosma, Maarten and Ichter, Brian and Xia, Fei and Chi, Ed H. and Le, Quoc V. and Zhou, Denny},
  booktitle = {Advances in Neural Information Processing Systems},
  volume    = {35},
  pages     = {24824--24837},
  year      = {2022},
  doi       = {10.52202/068431-1800},
  url       = {https://proceedings.neurips.cc/paper_files/paper/2022/hash/9d5609613524ecf4f15af0f7b31abca4-Abstract-Conference.html}
}

@inproceedings{wang2023selfconsistency,
  title     = {Self-Consistency Improves Chain of Thought Reasoning in Language Models},
  author    = {Wang, Xuezhi and Wei, Jason and Schuurmans, Dale and Le, Quoc V. and Chi, Ed H. and Narang, Sharan and Chowdhery, Aakanksha and Zhou, Denny},
  booktitle = {The Eleventh International Conference on Learning Representations},
  year      = {2023},
  url       = {https://openreview.net/forum?id=1PL1NIMMrw}
}

@article{yang2025qwen3,
  title         = {{Qwen3 Technical Report}},
  author        = {Yang, An and Li, Anfeng and Yang, Baosong and Zhang, Beichen and Hui, Binyuan and Zheng, Bo and Yu, Bowen and others},
  journal       = {arXiv preprint arXiv:2505.09388},
  year          = {2025},
  eprint        = {2505.09388},
  archivePrefix = {arXiv},
  primaryClass  = {cs.CL},
  doi           = {10.48550/arXiv.2505.09388},
  url           = {https://arxiv.org/abs/2505.09388}
}

@article{he2025deepmath,
  title         = {{DeepMath-103K}: A Large-Scale, Challenging, Decontaminated, and Verifiable Mathematical Dataset for Advancing Reasoning},
  author        = {He, Zhiwei and Liang, Tian and Xu, Jiahao and Liu, Qiuzhi and Chen, Xingyu and Wang, Yue and Song, Linfeng and Yu, Dian and Liang, Zhenwen and Wang, Wenxuan and Zhang, Zhuosheng and Wang, Rui and Tu, Zhaopeng and Mi, Haitao and Yu, Dong},
  journal       = {arXiv preprint arXiv:2504.11456},
  year          = {2025},
  eprint        = {2504.11456},
  archivePrefix = {arXiv},
  primaryClass  = {cs.CL},
  doi           = {10.48550/arXiv.2504.11456},
  url           = {https://arxiv.org/abs/2504.11456}
}

@inproceedings{xu2025kodcode,
  title     = {{KodCode}: A Diverse, Challenging, and Verifiable Synthetic Dataset for Coding},
  author    = {Xu, Zhangchen and Liu, Yang and Yin, Yueqin and Zhou, Mingyuan and Poovendran, Radha},
  booktitle = {Findings of the Association for Computational Linguistics: ACL 2025},
  pages     = {6980--7008},
  publisher = {Association for Computational Linguistics},
  address   = {Vienna, Austria},
  year      = {2025},
  doi       = {10.18653/v1/2025.findings-acl.365},
  url       = {https://aclanthology.org/2025.findings-acl.365/}
}

@inproceedings{llmdebate,
  title     = {Improving Factuality and Reasoning in Language Models through Multiagent Debate},
  author    = {Du, Yilun and Li, Shuang and Torralba, Antonio and Tenenbaum, Joshua B. and Mordatch, Igor},
  booktitle = {Proceedings of the 41st International Conference on Machine Learning},
  series    = {Proceedings of Machine Learning Research},
  volume    = {235},
  pages     = {11733--11763},
  publisher = {PMLR},
  year      = {2024},
  url       = {https://proceedings.mlr.press/v235/du24e.html}
}

@article{deepseekai2026deepseekv4,
  title   = {DeepSeek-V4: Towards Highly Efficient Million-Token Context Intelligence},
  author  = {{DeepSeek-AI}},
  journal = {arXiv preprint arXiv:2606.19348},
  year    = {2026}
}

@article{singh2025openai,
  title={Openai gpt-5 system card},
  author={Singh, Aaditya and Fry, Adam and Perelman, Adam and Tart, Adam and Ganesh, Adi and El-Kishky, Ahmed and McLaughlin, Aidan and Low, Aiden and Ostrow, AJ and Ananthram, Akhila and others},
  journal={arXiv preprint arXiv:2601.03267},
  year={2025}
}

@article{team2025internagent,
  title={InternAgent: When Agent Becomes the Scientist--Building Closed-Loop System from Hypothesis to Verification},
  author={Team, InternAgent and Zhang, Bo and Feng, Shiyang and Yan, Xiangchao and Yuan, Jiakang and Ma, Runmin and Hu, Yusong and Yu, Zhiyin and He, Xiaohan and Huang, Songtao and others},
  journal={arXiv preprint arXiv:2505.16938},
  year={2025}
}


\end{document}